\pdfoutput=1

\documentclass[11pt]{article}

\usepackage{acl}

\usepackage{times}
\usepackage{latexsym}

\usepackage[T1]{fontenc}

\usepackage[utf8]{inputenc}

\usepackage{microtype}

\usepackage{inconsolata}

\usepackage{graphicx}
\usepackage{multirow}
\usepackage{booktabs}
\usepackage{amsmath}
\usepackage[normalem]{ulem}
\usepackage{array}
\usepackage{todonotes}

\title{The Promises and Pitfalls of Using Language Models to Measure Instruction Quality in Education}

\author{Paiheng Xu \\
  University of Maryland \\
  \texttt{paiheng@cs.umd.edu} \\\And
  Jing Liu \\
  University of Maryland \\
  \texttt{jliu28@umd.edu} \\\And
  Nathan Jones \\
  Boston University \\
  \texttt{ndjones@bu.edu} \\\AND
  Julie Cohen \\
  University of Virginia \\
  \texttt{jjc7f@virginia.edu} \\ \And
  Wei Ai \\
  University of Maryland \\
  \texttt{aiwei@umd.edu}}

\begin{document}
\maketitle
\begin{abstract}
Assessing instruction quality is a fundamental component of any improvement efforts in the education system. However, traditional manual assessments are expensive, subjective, and heavily dependent on observers' expertise and idiosyncratic factors, preventing teachers from getting timely and frequent feedback. Different from prior research that mostly focuses on low-inference instructional practices on a singular basis, this paper presents the first study that leverages Natural Language Processing (NLP) techniques to assess multiple high-inference instructional practices in two distinct educational settings: in-person K-12 classrooms and simulated performance tasks for pre-service teachers. This is also the first study that applies NLP to measure a teaching practice that is widely acknowledged to be particularly effective for students with special needs. We confront two challenges inherent in NLP-based instructional analysis, including noisy and long input data and highly skewed distributions of human ratings. Our results suggest that pretrained Language Models (PLMs) demonstrate performances comparable to the agreement level of human raters for variables that are more discrete and require lower inference, but their efficacy diminishes with more complex teaching practices.  Interestingly, using only teachers' utterances as input yields strong results for student-centered variables, alleviating common concerns over the difficulty of collecting and transcribing high-quality student speech data in in-person teaching settings. Our findings highlight both the potential and the limitations of current NLP techniques in the education domain, opening avenues for further exploration.

\end{abstract}

\section{Introduction}
Evaluating instruction quality is a crucial component for most efforts aimed at improving an education system. Measuring teaching practices with validity and reliability can help advance the understanding of effective teaching strategies, enhance teacher professional learning, and thus improve students' academic performance \cite{adelman2003guide, wragg2011introduction, martinez2016classroom, desimone2017instructional}. Traditional methods to assess instruction quality require trained raters to observe classrooms based on established criteria. Despite its wide use in early childhood, K-12, and higher education, a long line of research has identified a range of issues with this human-based classroom observation approach, such as its lacking consistency across schools and different learning contexts due to time and resource constraints, human subjectivity, and varying levels of expertise among raters, among others \cite{wang-demszky-2023-chatgpt, kraft2018effect, kelly2020using}. 

In light of the limitations of human-based classroom observations, an emerging literature has started to use Natural Language Processing (NLP) techniques to automatically evaluate teaching practices by analyzing transcripts of classroom recordings \cite{demszky2023can, suresh2021using, nazaretsky2023empowering, demszky2021measuring, alic2022computationally, tan2023sit}. However, other than a few exceptions, these methods and related applications mostly focus on relatively low-inference instructional practices that rely on specific and concrete teaching moves \cite{rosenshine1971research}, such as 
the frequency of focusing (or open-ended) questions \cite{alic2022computationally}. In contrast, research is still scarce in exploring the feasibility of building automated measures on high-inference teaching practices, which require observers to draw conclusions and make interpretations based on many elements of teaching that are more aligned with existing classroom observation tools \cite{grossman2013measure}. One exception is \citet{wang-demszky-2023-chatgpt}, which uses zero-shot prompting to explore ChatGPT's ability in classifying classroom transcripts for a widely-used observation tool, Mathematical Quality Instruction (MQI) \cite{hill2008mathematical}. However, the authors find the performance of ChatGPT to be quite poor.

In this paper, we set out to evaluate to what extent state-of-the-art NLP methods can measure high-inference instructional practices. We conduct our study using data collected from both traditional in-person math classrooms in K-12 schools and simulated performance tasks designed for pre-service teachers. Our study is particularly novel in its inclusion of teaching practices that are widely acknowledged to be effective for students with special needs in general math classrooms \cite{fuchs2008intensive, gersten2009mathematics}. While research has been progressing quickly in the field of using NLP to measure teaching practices, to our knowledge, this is the first study that incorporates measures that target students with special needs \cite{wang2024artificial}.
For general math education, we evaluate 11 observation variables across four dimensions of math instruction quality in in-person classrooms from MQI. 
Comparing the performance of pretrained language models (PLMs) for multiple high-inference teaching practices allows us to evaluate the relative performance of language models based on the nature of measures. 

Our analysis shows that the performances of fine-tuned PLMs on measuring high-inference teaching practices vary by the levels of required pedagogical expertise. PLMs work decently well for variables largely depending on lexical usage such as mathematical language richness, achieving agreement levels on par with human experts. In contrast, PLMs demonstrate worse performances on variables that require further inferences that are also challenging for human experts, such as the precision of mathematical language and clarity of mathematical content. 

We identify and address several challenges that negatively impact the performance of PLMs. The first challenge is the highly skewed distribution of labels due to the lack of high-rating teaching samples in practice. Across both datasets, high-rating samples consist of less than $10\%$ of the dataset for the majority of variables. We address this issue by adopting class-weighted loss to help predict some variables, but the performance gain is limited. Another challenge comes from the length of transcripts used to infer teaching practices, which are typically from 5- to 15-minute classroom recordings. 
Such noisy and long inputs impede PLMs from focusing on the relevant information.
We adopt a two-stage strategy by first predicting relevant sentences and then making the model focus on predicted relevant sentences. We find this strategy benefits variables 
whose assessment relies on multiple sentences. Thus, even if the relevance classifier falsely rules out some relevant sentences, there are still cues for PLMs to make predictions.
To this end, we also use ChatGPT for zero-shot aspect summarization as the relevance model in the first stage but the extracted sentences are mostly irrelevant according to the rubrics, which is consistent with \citet{wang-demszky-2023-chatgpt}.

Although it is preferable to use all dialogue in a classroom for measuring teaching practices, in traditional classroom settings, the collection of student speech data is often challenging due to microphone setups and consent requirements. Surprisingly, using teachers' utterances alone achieves reasonable performances for student-oriented observation variables. 
This finding underscores the potential of focusing on teacher-led discourse for comprehensive classroom assessment.

We summarize our contributions as follows. (1) We present to-date the most comprehensive study of assessing high-inference instruction quality using PLMs, encompassing both in-person and online educational settings, with a wide coverage of observation variables notably including those deemed to be important for special education. (2) We mitigate two challenges inherent to this task, stemming from its real-world setup and constraints. We introduce a two-stage strategy to mitigate the impact of long and noisy input and employ a class-weighted loss to compensate for the lack of high-rating teaching samples. (3) We demonstrate several practical implications when using PLMs to measure teaching practices, including the dependence of model performances on the required pedagogical expertise and the effectiveness of using only teachers' utterances for capturing classroom observations, even for student-centered variables.

\section{Data}
\label{sec:data}
\begin{table*}[t]
\centering
\scriptsize
\begin{tabular}{>{\hspace{0pt}}m{0.08\linewidth}>{\hspace{0pt}}m{0.175\linewidth}>{\hspace{0pt}}m{0.643\linewidth}} \toprule
\multicolumn{1}{>{\hspace{0pt}}m{0.08\linewidth}}{Variable} & \multicolumn{1}{>{\hspace{0pt}}m{0.175\linewidth}}{Definition}                                                        & \multicolumn{1}{>{\arraybackslash\hspace{0pt}}m{0.643\linewidth}}{High-Rating Examples}\\ \midrule
Unpacking                                                              & A statement making sense of the context, quantities, or mathematical relationships.                & - {\it ``That word `about' means I do not need to know exactly how much – I need to estimate. And to estimate, I can find close, friendly numbers.''} (Example of strategically addressing concepts that may be confusing for students. Other examples may address vocab, context, using tables, etc.)                                                                                                                                                                                                 \\ \midrule
Self-Instruction                                                       & An explicit statement made by the teacher which references their thinking and/or processes related to making sense of problems. & - {\it ``I’m going to ask myself, what is this problem about?''} (Explicit general self-instruction example, such examples should be paired with specific self-instruction statements or justifications to be rated as high.)\par\vspace{0.3em}\par{}- {\it ``I know how many apples and how many bananas Naveah has in her basket.''} (Explicit specific self-instruction example).\par\vspace{0.3em}\par{}- {\it ``What is this problem asking me? By putting it in my own words, I can make sure I get what’s going on.''} (This justification makes clear to students why putting a problem in their own words can help them make sense of the problem)  \\ \midrule
Self-Regulation                                                        & A self-instruction statement that supports students' emotional regulation.                                                      & - {\it ``I just read this information and thought, this is a lot of information. When I’m feeling overwhelmed by the amount of information in a problem, I know I can go back and re-read the problem sentence by sentence to help me break the problem down into smaller, more manageable chunks.''} (Example of explicit reference to a self-regulation strategy and reference to how it can be applied across problems more broadly)                                                                 \\ \bottomrule
\end{tabular}
\caption{Rating rubrics of selected metacognitive modeling components in the SimSE dataset.}
\label{tab:example_simse}
\end{table*}

\begin{table*}
\scriptsize
\centering
\begin{tabular}{>{\hspace{0pt}}m{0.14\linewidth}>{\hspace{0pt}}m{0.37\linewidth}>{\hspace{0pt}}m{0.40\linewidth}} \toprule
\multicolumn{1}{>{\hspace{0pt}}m{0.14\linewidth}}{Dimension} & \multicolumn{1}{>{\hspace{0pt}}m{0.37\linewidth}}{Variable Definition} & \multicolumn{1}{>{\arraybackslash\hspace{0pt}}m{0.40\linewidth}}{High-Rating Examples/Criteria}                                                                                                                                                                                                   \\ \midrule
Richness of the Mathematics                                                                                  & \textit{Mathematical Language (MLANG)} captures how fluently the teacher (and students) use mathematical language and whether the teacher supports students’ use of mathematical language.                                                                                                                         & 1. Density of mathematical language is high during periods of teacher talk.\par{}2. Moderate density, but also explicitness about terminology, reminding students of meaning, pressing students for accurate use of terms, encouraging student use of mathematical language.                                     \\ \midrule
Working with Students and Mathematics                                                                        & \textit{Remediation of Student Errors and Difficulties (REMED)} records instances of remediation in which student misconceptions and difficulties with the content are addressed. \textit{Conceptual remediation} gets at the root of student misunderstandings, rather than repairing just the procedure or fact. & - {\it “I noticed that some of you forgot to multiply both sides of the equation by x. What happens if you multiply one side by x and not the other?”} [The class continues to discuss at length why you need to multiply on both sides.] (Teacher engages in conceptual remediation systematically and at length. )  \\ \midrule
Errors and Imprecision                                                                                    & {\it Imprecision in Language or Notation (LANGIMP)} captures problematic uses of mathematical language or notation                                                                                                                                                                      & Imprecision occurs in most or all of the segment, OR imprecision obscures the mathematics of the segment.                                                                                                                                                                                               \\ \midrule
Student Participation in Meaning-Making and Reasoning                                                                                 & {\it Students Provide Explanations (STEXPL)} captures when students provide a mathematical explanation for an idea, procedure, or solution.                                                                                                                                                                              & Student explanations characterize much of the segment.                                                                                                                                                                                                                                                          \\ \bottomrule
\end{tabular}
\caption{Rating rubrics of selected MQI variables in the NCTE dataset.}
\label{tab:example_ncte}
\vspace{-0.5em}
\end{table*}

To comprehensively evaluate PLMs' ability to evaluate instruction quality, we select two datasets--the Simulation for Special Education ({\bf SimSE}) dataset and the National Center for Teacher Effectiveness ({\bf NCTE}) Transcript dataset \cite{demszky-hill-2023-ncte}, which capture both simulated teaching sessions for pre-service teachers and in-person K-12 classrooms. We introduce the details of these two datasets and their corresponding instruction quality measurement.

\subsection{SimSE -- Metacognitive Modeling}
The SimSE dataset is collected as part of TeachSim,\footnote{https://www.teachsim.org/} a teaching simulation platform for pre-service teachers. 
All participating teachers complete 5-minute simulation sessions and receive coaching on how to better elicit students’ thinking. The goal is to improve teachers' use of metacognitive modeling, a teaching practice that involves the narration of actions, decisions, and thought processes while demonstrating metacognitive strategies \cite{archer2010explicit, montague1992effects}. The dataset in our analysis focuses on students with special needs in general math education, collected from Fall 2022 to Spring 2023. The simulation sessions in the dataset focus on teaching metacognitive modeling while unpacking (not solving) a word problem.
\citet{wilson2016teaching, mcleskey2017high} show that such a strategy is one of the most critical practices to support students with special needs.

Specifically, metacognitive modeling contains five components, among which {\it Unpacking}, {\it Self-Instruction}, and {\it Self-Regulation} are the most important. Table \ref{tab:example_simse} shows the definitions and high-rating examples for these components. 
Rating these high-inference components requires drawing conclusions from and interpreting several elements of teaching. For example, {\it Unpacking} can be applied to address context, quantities, and mathematical relationships in word problems. The core idea is that teachers should make sense of the concepts strategically so that the understanding can be generalized across problems.
{\it Self-regulation} mainly concerns students' emotional regulation and such statements often contain emotion-related words such as ``overwhelmed'' and ``confused'', though one can also implicitly express such emotions by saying ``there is a lot of information in the problem''. More importantly, successful demonstration of self-regulation requires the teacher to accompany emotion expression by regulation strategies and how these strategies support sense-making in general.
{\it Self-instruction} is the most difficult component to identify as it requires multiple types of evidence. Explicit, general self-instruction statements should be paired with specific statements or justifications to support the understanding of the current problem and metacognitive thinking for future problem-solving. 

The other two components of metacognitive modeling require less pedagogical expertise: \textit{Objective} evaluates the quality of the teacher stating the two goals: paying attention to the teacher’s thinking (so that students may use it in their future problem-solving) and making sense of the problem. 
\textit{Ending} requires teachers to refrain from solving the problem or presenting strategies or operations.

All these components are scored as low, mid, and high (1 to 3) by experts, where the mid rating applies to samples that partially meet the criteria. 
The three-way rating scale was developed by instruction experts to balance granularity and reliability while providing meaningful differentiation of teaching quality. While a more granular scale might provide more nuanced insights, it would also exacerbate the difficulty of achieving a high inter-rater reliability, which has been demonstrated as a difficult task in classroom observations \cite{ho2013reliability}.
We note that collecting such educational data poses significant challenges, notably the high cost of annotation and the necessity for well-defined rubrics that delineate each category unambiguously. 

\begin{figure}
    \centering
    \includegraphics[width=1\linewidth]{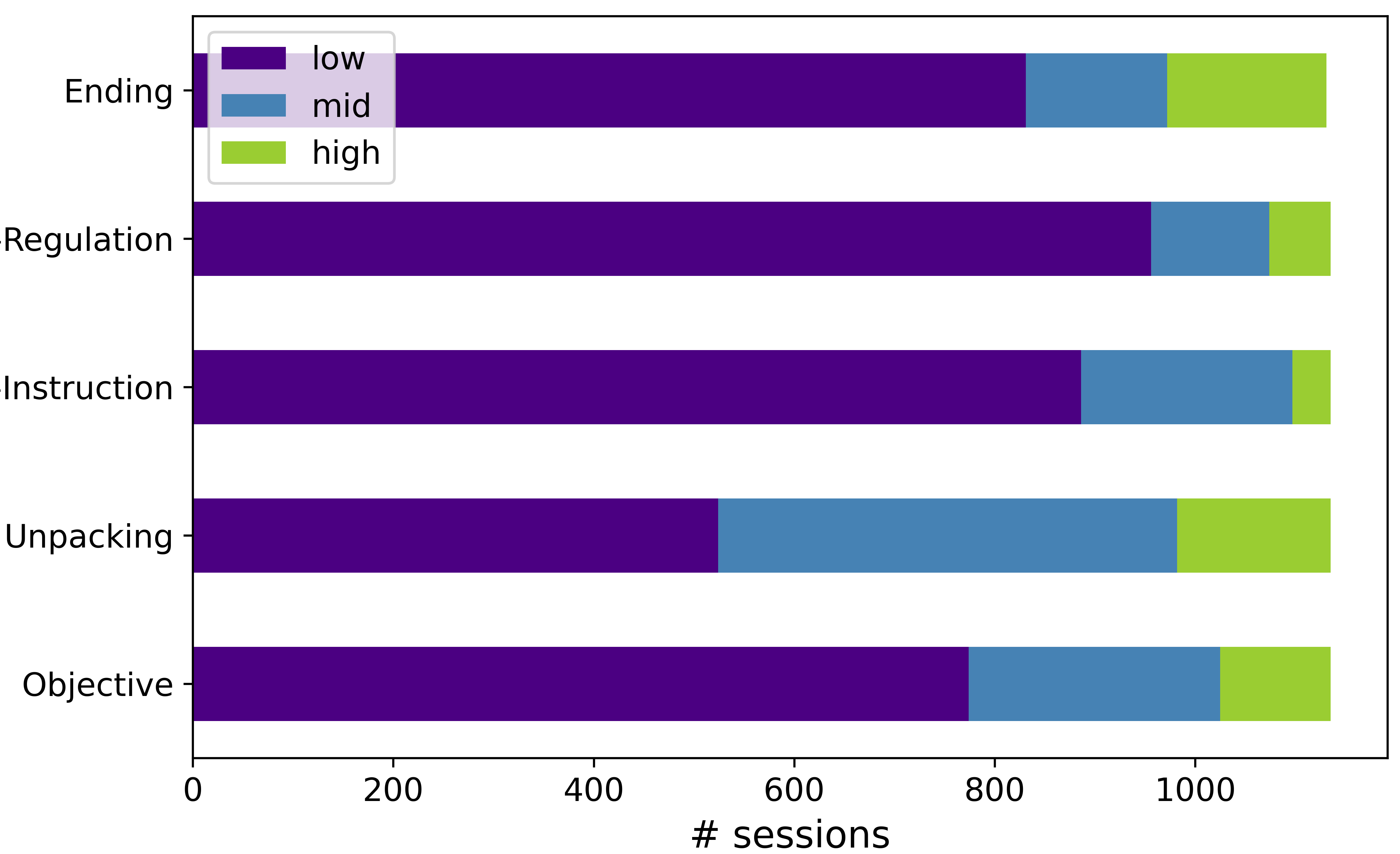}
    \caption{Label distribution for metacognitive modeling components in the SimSE dataset.}
    \label{fig:simse_dist}
\end{figure}

\paragraph{Dataset Statistics.} The SimSE dataset contains 1,135 five-minute transcribed teaching sessions (teachers' talk only), each annotated with the five metacognitive modeling components. On average, each transcript has 416.1 words (std: 187.9). A subset of transcripts are annotated by two or three raters. The Spearman correlations between raters are shown in Table \ref{tab:simse_test} (detailed computation process in Appendix~\ref{app:rater}). 
We round the average score for each session as the prediction label.
The label distributions are highly skewed towards the low ratings, which may account for up to $80\%$ of the samples for some components, as shown in Figure~\ref{fig:simse_dist}. 
There might be multiple reasons why high-rating teaching samples are rare. For example, pre-service teachers have not yet worked in real classrooms and are still developing their teaching skills.

\begin{figure}
    \centering
    \includegraphics[width=1\linewidth]{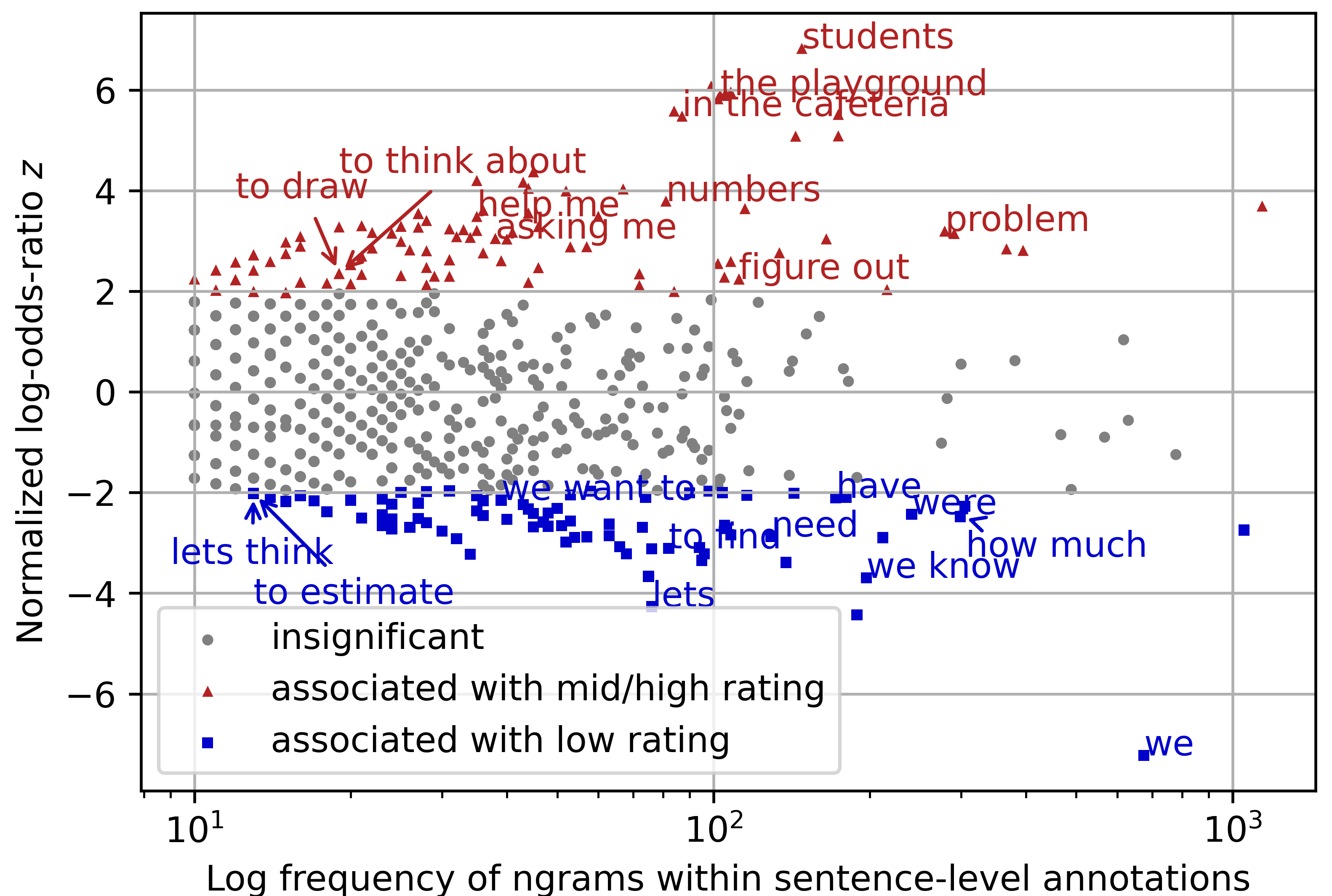}
    \caption{Top $n$-grams ($n \in \{1, 2 ,3\}$) from sentences that are annotated relevant to the mid-high and low ratings for Self-Instruction. Only $n$-grams with $95\%$ confidence level are colored ($z>1.96$). }
    \label{fig:lexical}
    \vspace{-0.5em}
\end{figure}

\paragraph{Lexical Analysis.}
We examine the linguistic properties of teaching language by analyzing the lexical differences between low and high rating samples. A subset of 199 transcripts have sentence-level annotations of their relevance to each of the five components. The raters were asked to select the evidence sentences that helped them score each component following the rubrics.
We follow \citet{monroe2008fightin} and use normalized log-odds-ratio $z$ to identify the n-grams more associated with each rating level. Given the lack of high ratings, we combine mid and high ratings and compare them with low-rating transcripts. 

From the visualization for the Self-Instruction component in Figure \ref{fig:lexical}, phrases like ``help me'' can be signals for justification statements, which are one of the required evidence for mid/high ratings. However, we observe conflicting signals for the other type of required evidence, specific self-instruction statements. Typical self-instruction phrases such as  ``think about'', ``to draw'', and ``figure out'' are associated with mid/high ratings while similar phrases such as ``lets think'', ``to find'', and ``to estimate'' are associated with low ratings. Therefore, the identification of mid/high rating components requires high-inference ability and pedagogical expertise beyond lexical cues, highlighting the difficulty of automatic measurement.

The SimSE dataset focuses on students with special needs in simulated teaching sessions. We posit that it represents a simpler context for analysis compared to the in-person classroom setting, regardless of the complexity of the variables being examined. This is because the setting of the SimSE dataset is similar to modular online teaching sessions which are relatively short and comprise teacher utterances only. We complement the SimSE dataset by using the NCTE dataset, which is collected from in-person, math classrooms in K-12 education.

\subsection{NCTE -- MQI}
The NCTE dataset is the largest publicly available dataset of U.S. classroom transcripts linked with classroom observation scores. 
The raters evaluate math teaching quality based on MQI \cite{hill2008mathematical} instruments along four dimensions: {\it Richness of the Mathematics, Working with Students and Mathematics, Errors and Imprecision,} and {\it Student Participation in Meaning-Making and Reasoning}. Each dimension has multiple observation variables.
We experiment on a total of eleven variables and present one variable per dimension for presentation purposes, as shown in Table \ref{tab:example_ncte}.
The description of the remaining seven variables can be found in Appendix \ref{app:mqi_additional}. 
Similar to the SimeSE dataset, these variables are scored on a scale of 1-3, corresponding to low, mid, and high ratings.

These variables also vary in the amount of information and inferences required for human evaluation. For example, {\it Mathematical Language} (MLANG) is relatively straightforward as it mainly depends on the density of mathematical language usage, which is readily identifiable once a dictionary of such words is available. In contrast, {\it Remediation of Student Errors and Difficulties} (REMED) and {\it Imprecision in Language or Notation} (LANGIMP) require further inferences and information beyond the mathematical language used in classrooms. For example,
high ratings of REMED should remediate student misunderstanding at a conceptual level. Such remediation behavior and relevant discussion should also occupy a large proportion of the segment used for annotation to be considered as ``high''. 
LANGIMP covers math content, including notation and languages used to convey technical terms. Its evaluation requires an understanding of these contents and the ability to identify when they are imprecisely expressed.
{\it Students Provide Explanations} (STEXPL) is a student-oriented variable. 

\paragraph{Dataset Statistics.}
The NCTE dataset consists of 1,660 4th and 5th grade elementary mathematics observations collected by NCTE between 2010-2013. The transcripts are anonymized and represent data from 317 teachers across 4 school districts that serve largely historically marginalized students. 
The original dataset has transcripts for the whole class, which are typically 45-60 minutes long, while the MQI instrument observation variables apply to 7.5-minute segments.
Therefore, we align the transcripts with MQI observations, resulting in 9,886 segments, each rated on 11 MQI variables. 

Similar to the SimSE dataset, the label distribution in the NCTE dataset is also highly imbalanced. High-rating segments are extremely rare and low-rating segments constitute large proportions for most variables, which reverberates the practical challenge of lacking highly rated teaching activities in real-world education data.

Additionally, since the collection of student talk is challenging in practice, student utterances are often inaudible. 
We experimented with using teacher utterances only and including student utterances as input text. 
For the first input setting, we concatenate all teacher utterances as input. To include student utterances, we format the input in the transcript style, where each utterance is \texttt{<speaker>:<utterance>}. \texttt{<speaker>} can be either the teacher, a student, or multiple students. All speakers are anonymized, and the students have identifiers (e.g., \texttt{student A}) generated by professional transcribers when possible. We discuss the results of these two formats in Section \ref{sec:results}.
On average, each segment contains 647.6 words (std: 230.2) when using teacher utterances only, and 750.6 words (std: 259.9) when including student utterances. 

\begin{figure}[t]
    \centering
    \includegraphics[width=1\linewidth]{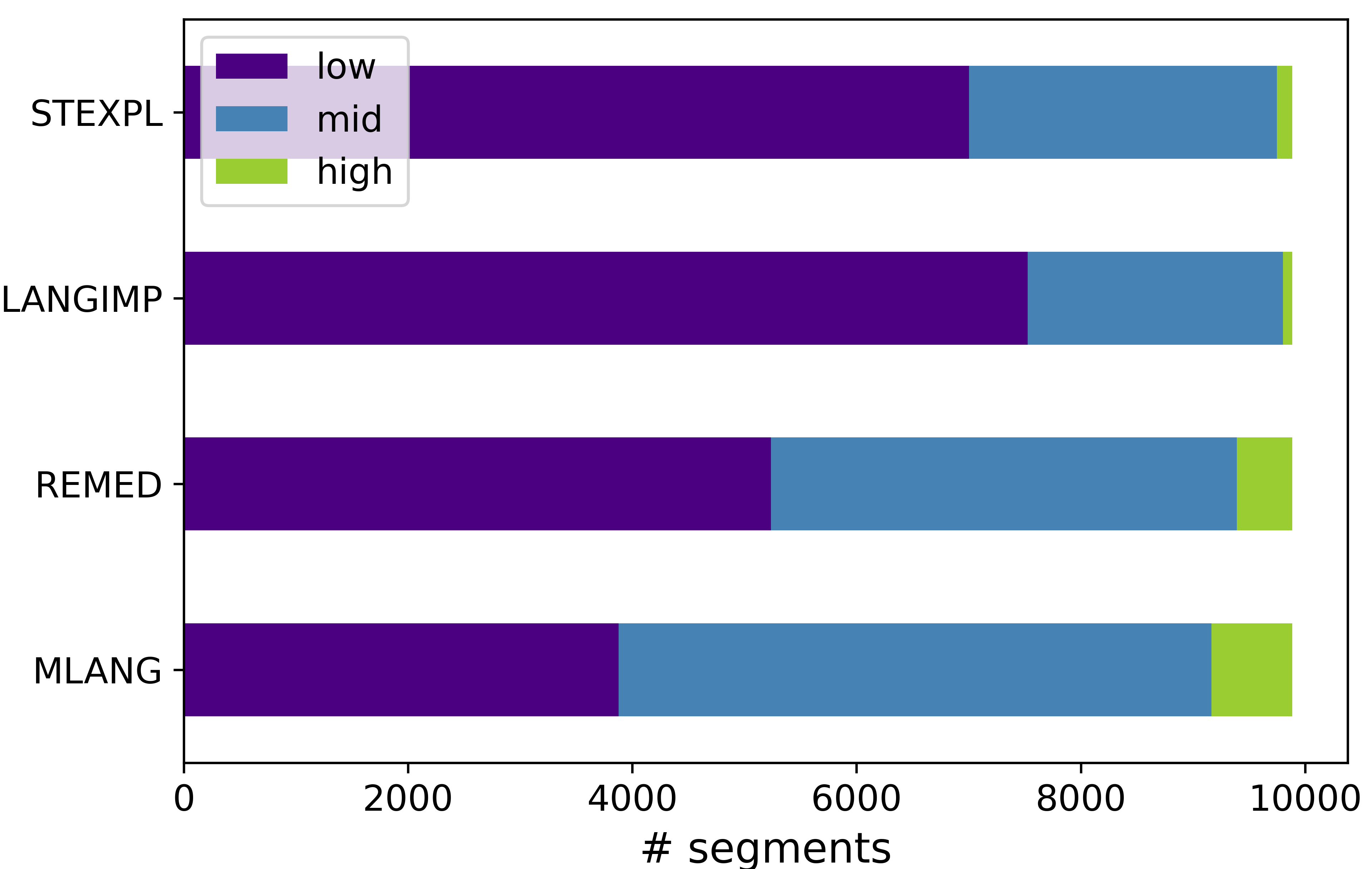}
    \caption{Label distribution for MQI variables in the NCTE dataset.}
    \label{fig:ncte_dist}
    \vspace{-0.5em}
\end{figure}

Despite the differences in variable definitions and educational settings between metacognitive modeling in the SimSE dataset and MQI variables in the NCTE dataset, assessing the quality of these variables in both datasets requires raters to have strong domain knowledge and the ability to draw conclusions and interpret several elements of teaching.
The SimSE dataset is tailored towards metacognitive modeling for students with special needs, and the NCTE dataset concentrates on the mathematical instruction quality. 
Therefore, we hope our study provides insights into how effectively PLMs can measure high-inference and highly domain-specific instructional practices in both simulated teaching sessions and in-person classrooms, two common educational settings currently.

\section{Methods}
\label{sec:methods}
Measuring instructional practices is naturally formulated as text classification tasks. To comprehensively understand language models' ability in these tasks, we experiment with a wide range of PLMs. Specifically, we fine-tuned the following models by adding a classification head on top of the hidden layers, using the implementation from Transformer \cite{wolf2020transformers}, i.e., BERT~\cite{devlin2018bert}, DistilBERT~\cite{sanh2019distilbert}, XLNet~\cite{yang2019xlnet}, and RoBERTa~\cite{liu2019roberta}.
We also fine-tuned Llama2-7B~\cite{touvron2023llama} with QLoRA~\cite{dettmers2024qlora} in the instruction-following style by converting the dataset to a next-word prediction setup. More discussions on model choices are in Appendix~\ref{app:training}.

We identify two primary challenges in automatically measuring teaching quality: (1) the text input is long and noisy, making it hard for PLMs to focus on the relevant sentences for each variable; and (2) the label distributions are highly imbalanced, as there are few high-rating teaching transcripts following the standards in the rubric for each observation variable. We discuss how we mitigate these two challenges with available resources next.

\subsection{Challenge of Long Input}
\label{sec:long_input}
The average lengths for simulation sessions and classroom segments in the SimSE and NCTE datasets are $416.1$ and $647.6$ words, respectively. Depending on the definitions of the observation variables, the amount of irrelevant text in the model input varies. As shown in Table \ref{tab:simse_sentence} for the SimSE dataset, components like Objective, Self-Regulation, and Ending have fewer than or close to two relevant sentences per simulation session, while Self-Instruction has 8 and Unpacking has 21. 

To reduce the impact of irrelevant information, we develop a two-stage prediction strategy that first predicts the relevance of each sentence and then scores on the concatenation of sentences predicted as relevant. 
Formally, for each component, we first build a relevance model $\mathcal{Y}_R(s_i)$ that predicts the relevance of each sentence $s_i$ in transcript $T$.
We then train a classification model $\mathcal{Y}_S(S_R)$ whose input concatenates the sentences predicted as relevant by $\mathcal{Y}_R$, i.e., $S_R = \{\mathcal{Y}_R(s_i)=1 | s_i \in T\}$.

Additionally, given the recent success of ChatGPT in summarization tasks~\cite{zhang2024benchmarking}, especially aspect summarization for conversations~\cite{yang2023exploring}, we also experiment with using ChatGPT as the relevance model in the first stage.
The two-stage model can then be defined as $\mathcal{Y}_S(\mathcal{S}(T))$, where $\mathcal{S}(\cdot)$ is the summarization model, and its output is directly used as the input for the classification model in the second stage.

\subsection{Challenge of Label Imbalance}
\label{sec:imbalance}
High-rating transcripts are rare in our datasets and in real-world classrooms/simulation sessions as shown in the largest study on teaching quality to date by \citet{kane2012gathering}.
Therefore, differentiating mid- and high-rating transcripts from low-rating ones is helpful in practice. We group the mid- and high-rating transcripts and convert the three-way classification task to a binary one. We report the performances in both settings.

Moreover, the low-rating transcripts are the major class for almost all observation variables, taking up from $53\%$ to $89\%$ of the samples. 
For both binary and three-way settings, we use cross-entropy loss weighted by the inverse class frequency, defined as $\mathcal{L}(y, \hat{y}) = - \sum_{j=1}^{c} w_j y_j \log(\hat{y_j})$,
where $y_j$ = 1 if training sample $i$ belongs to class $j$ and
0 otherwise. $y$ and $\hat{y}$ are the ground-truth label and predicted label, respectively. $c$ is the number of classes, and $w_j = \frac{N}{c \times N_j}$ is the weight for class $j$, where $N$ and $N_j$ are the total number of training examples and the number of class $j$ respectively.

\section{Experiments}
\label{sec:experiments}
To evaluate the models and techniques introduced above, we randomly split the simulation sessions and classroom segments from the SimSE and NCTE datasets into train/dev/test sets with a ratio of $60\%$/$20\%$/$20\%$. For each observation variable, we fine-tune the models mentioned in Section~\ref{sec:methods} on the training set, including variants that apply the two-stage prediction strategy introduced in Section~\ref{sec:long_input} (only on the SimSE data due to the availability of sentence-level annotations) and the ones that use the weighted loss defined in Section~\ref{sec:imbalance}. 

We select the best model combination based on dev set performances and report their performances on the test set in Table \ref{tab:simse_test} and Table \ref{tab:ncte_test} for the SimSE and NCTE datasets respectively.
We follow this procedure for three-way and binary settings and report macro-F1 as the evaluation metric for both. 
We use Majority F1 as the baseline of predicting the majority rating to show the performance gain for each variable.
For the three-way setting, we also report the Spearman correlation, a non-parametric measure of ranking correlation, to account for the ranking of the labels (i.e., low, mid, and high).
We repeat experiments for each observation variable five times and report the mean and standard deviation of the performance scores.

\begin{table}
\centering
\resizebox{\linewidth}{!}{%
\begin{tabular}{lrrrrrr} \toprule
\multirow{2}{*}{Component} & \multirow{2}{*}{$\rho_\text{rater}$} & \multicolumn{3}{c}{Three-way}      & \multicolumn{2}{c}{Binary}  \\
\cmidrule(lr){3-5}\cmidrule(l){6-7}
                           &                              & Majority F1 & F1        & Spearman  & Majorify F1 & F1            \\ \midrule
Objective                  & 0.79                         & 27±1   & 75±3 & 0.83±0.02 & 41±1   & 91±2     \\
Unpacking                  & 0.57                        & 22±1   & 64±1 & 0.61±0.03 & 35±2   & 77±2     \\
Self-Instruction           & 0.34                         & 29±0   & 60±7 & 0.56±0.04 & 43±0   & 76±2     \\
Self-Regulation            & 0.62                         & 31±0   & 65±4 & 0.68±0.05 & 46±0   & 85±4     \\
Ending                     & 0.78                         & 28±1   & 81±3 & 0.90±0.01 & 43±0    & 95±1     \\ \bottomrule
\end{tabular}
}
\caption{Test set performance of selected models for metacognitive modeling components in the SimSE dataset. $\rho_\text{rater}$ shows the weighted average Spearman correlation among human raters in the three-way setting. F1 scores are in percentage. All Spearman correlations on the test sets are significant with 95\% confidence.}
\label{tab:simse_test}
\end{table}

\begin{table}
\centering
\resizebox{\linewidth}{!}{%
\begin{tabular}{lrrrrr} \toprule
\multirow{2}{*}{Variable} & \multicolumn{3}{c}{Three-way}      & \multicolumn{2}{c}{Binary}  \\
\cmidrule(lr){2-4}\cmidrule(l){5-6}
                      & Majority F1 & F1         & Spearman  & Majority F1 & F1            \\ \midrule
MLANG                 & 23±0   & 53±1 & 0.42±0.02 & 38±0   & 69±1     \\
REMED                 & 23±0   & 43±1 & 0.28±0.01 & 35±0   & 65±1     \\
LANGIMP               & 29±0   & 44±4 & 0.20±0.04 & 43±0   & 61±1     \\
STEXPL                & 28±0   & 47±2 & 0.35±0.01 & 42±0   & 69±1     \\ \bottomrule
\end{tabular}
}
\caption{Test set performance of selected models for the four selected MQI variables in the NCTE dataset. Inputs use teacher text only. F1 scores are in percentage. All Spearman correlations on the test sets are significant with 95\% confidence.}
\label{tab:ncte_test}
\vspace{-0.5em}
\end{table}

\begin{table}
\centering
\small
\begin{tabular}{lrrr} \toprule
\multirow{2}{*}{Variable} & \multicolumn{2}{c}{Three-way} & Binary     \\
\cmidrule(lr){2-3}\cmidrule(lr){4-4}
                      & F1 (\%)       & Spearman           & F1 (\%)         \\ \midrule
MLANG                 & 52±1 & 0.40±0.02          & 68±2  \\
REMED                 & 44±1 & 0.27±0.02          & 63±2  \\
LANGIMP               & 42±4 & 0.19±0.02          & 60±1  \\
STEXPL                & 48±3 & 0.37±0.02          & 68±1  \\ \bottomrule
\end{tabular}
\caption{Test set performance of selected models for the four selected MQI variables in the NCTE dataset. Inputs are in the transcript format. All Spearman correlations
on the test sets are significant with 95\% confidence}
\label{tab:ncte_test_transcript}
\end{table}



\begin{table}[t]
\centering
\resizebox{\linewidth}{!}{%
\begin{tabular}{lrrrr} \toprule
Component        & Avg. $\#$ Sent. & Majority F1 & F1     & Positive F1  \\ \midrule
Objective        &    1.9        & 49±0       & 88±1 & 78±3          \\
Unpacking        &    21.9        & 38±0      & 79±1 & 84±1       \\
Self-Instruction &     8.4       & 43±0      & 75±1  & 62±2       \\
Self-Regulation  &    0.7        & 50±0      & 78±2 & 57±5         \\
Ending           &     2.3       & 48±0      & 65±2 & 35±4       \\ \bottomrule
\end{tabular}
}
\caption{Sentence prediction results for SimSE dataset. Avg. $\#$ Sent. is the average number of relevant sentences in each simulation session. Positive F1 is the F1 score of the positive class (relevant sentences). F1 scores are in percentage.}
\label{tab:simse_sentence}
\vspace{-0.5em}
\end{table}

\begin{table*}
\centering
\resizebox{\linewidth}{!}{%
\begin{tabular}{lccccccccccccccccc} \toprule
\multirow{2}{*}{Component} & \multicolumn{5}{c}{Normal}                   & \multicolumn{4}{c}{Weighted}        & \multicolumn{4}{c}{Two-stage}       & \multicolumn{4}{c}{Two-stage + Weighted}  \\

\cmidrule(lr){2-6}\cmidrule(lr){7-10}\cmidrule(lr){11-14}\cmidrule(lr){15-18}
                           & distilbert & bert  & xlnet & roberta & Llama2 & distilbert & bert  & xlnet & roberta & distilbert & bert  & xlnet & roberta & distilbert & bert  & xlnet & roberta      \\ \midrule
Objective                  & \textbf{74±3} & 67±13 & 69±15          & 69±9          & 41±7   & 73±2          & 56±21 & 55±26 & 65±22         & 71±4       & 72±4  & 50±23 & 62±12   & 70±2          & 61±12 & 56±12 & 64±12     \\
Unpacking                  & 64±5          & 49±24 & 57±20          & \textbf{67±4} & 44±9   & 67±5          & 59±12 & 41±27 & \textbf{67±4} & 63±6       & 63±6  & 56±20 & 65±9    & 65±4          & 64±4  & 38±20 & 57±20     \\
Self-Instruction           & 63±10         & 57±11 & 55±16          & 48±14         & 36±17  & \textbf{63±9} & 56±18 & 40±16 & 44±18         & 52±6       & 55±16 & 41±19 & 48±15   & 60±10         & 47±12 & 35±9  & 42±13     \\
Self-Regulation            & 50±5          & 45±11 & 47±11          & 42±16         & 31±0   & 53±9          & 42±15 & 39±9  & 45±19         & 65±5       & 61±8  & 49±14 & 42±11   & \textbf{69±5} & 66±9  & 56±8  & 60±8      \\
Ending                     & 69±4          & 60±18 & \textbf{76±11} & 66±7          & 44±10  & 67±4          & 51±22 & 72±24 & 51±17         & 57±9       & 56±10 & 50±11 & 55±14   & 61±6          & 61±7  & 40±21 & 61±10     \\ \bottomrule
\end{tabular}
}
\caption{The F1 scores (\%) of predicting metacognitive modeling components on the Dev set of SimSE dataset in the three-way setting. The best-performing models are bolded.}
\label{tab:simse_dev_three}
\end{table*}

The test set results for selected models are shown in Table~\ref{tab:simse_test} and Table~\ref{tab:ncte_test} for the SimSE and NCTE datasets, respectively.
Results for the other seven MQI variables are shown in Appendix~\ref{app:mqi_additional}, and hyperparameter searching and other training details are described in Appendix~\ref{app:training}.

\section{Results and Discussion}
\label{sec:results}
In general, models perform much better on metacognitive modeling components than MQI variables, partly due to the shorter input length and the teacher-only setup.
However, we observe considerable variations across the variables.
We discuss the insights in detail as follows.

\paragraph{LLMs work better for variables that require less pedagogical expertise.}
In Table~\ref{tab:simse_test}, Objective and Ending have significant Spearman correlations over $0.8$ in the three-way setting and F1 scores over $0.9$ in the binary setting.
However, components that require higher levels of pedagogical expertise, i.e., Unpacking, Self-Instruction, and Self-Regulation, have notably worse performance, although still obtaining comparable correlation levels as achieved by human raters.
Similarly, in Table~\ref{tab:ncte_test}, LLMs work better for MLANG which can be assessed by lexical usage, while perform worse on REMED and LANGIMP, which require further inferences of mathematical knowledge.
Similar trends can be observed for other MQI variables in Table~\ref{tab:ncte_test_additional}.

\paragraph{Using only teacher utterances as input captures most observations, even for student-oriented variables.}
For NCTE, we experimented with using only teacher utterances and including student utterances (Table~\ref{tab:ncte_test} and Table~\ref{tab:ncte_test_transcript}, additional results in Appendix \ref{app:mqi_additional}).
We find that using only teacher utterances achieves comparable and even better results for all MQI variables, despite ignoring student utterances.
This is particularly interesting for the student-oriented variables (e.g., STEXPL) in the Student Participation in Meaning-Making and Reasoning dimension of MQI.
Student talk is hard to collect in practice since it requires setting up multiple mics across the classroom and obtaining privacy consent. The fact that we can measure student activities with teacher utterances alone underscores the potential of focusing on teacher-led discourse for comprehensive classroom assessment.

The results above are from models selected from various model types and combined with different techniques described in Section~\ref{sec:methods}.
We next present the results and discuss their implications.

\paragraph{Class-weighted loss marginally improves the performances.}
Variables that have more balanced class distributions tend to have better performance, but the intrinsic difficulty of the variables seems to be the driving factor.
For example, Self-regulation has a more skewed distribution but worse performance compared to Self-Regulation.
Similar observations can be made when comparing Unpacking with Objective or Ending.
Moreover, the adoption of class-weighted loss achieved best performance on 10 out of the 16 variables in the three-way setting (Tables~\ref{tab:simse_dev_three} and \ref{tab:ncte_dev_three}) and 9 out of 16 in the binary setting (Tables~\ref{tab:simse_dev_binary} and \ref{tab:ncte_dev_binary}), although the differences are small.
Therefore, class-weighted loss helps but with a marginal effect.

\subsection{Discussion}
\paragraph{Dealing with long input.}
We employ a two-stage prediction strategy, as described in Section~\ref{sec:long_input}.
Due to data availability, we only have sentence-level annotations on a subset of the SimSE dataset.
We apply the same experiment setup for the classification task on this sentence-level relevance prediction task.
The results are shown in Table~\ref{tab:simse_sentence}.
From Tables~\ref{tab:simse_dev_three} and \ref{tab:simse_dev_binary}, we find that the two-stage strategy helps Self-Instruction and Self-Regulation in the three-way setting,
despite their moderate F1 scores on the relevant (positive) class in the first stage.
We suspect that components with multiple relevant sentences (i.e., Self-Instruction) per session are less prone to the error of the relevance classifier. That is, missing one or two relevant sentences would result in empty inputs for Objective and Ending, which usually have only a few relevant sentences per session.
For Self-Regulation, as this teaching practice depends on emotion-related expressions, we conjecture that the predicted relevant sentences are likely from mid- and high-rating samples.

We also use ChatGPT as a zero-shot summarization model to avoid reliance on costly sentence-level annotations.
However, our preliminary analysis shows that the relevant utterances retrieved by ChatGPT are not helpful.
For example, as shown in Figure~\ref{fig:prompt}, when prompted to summarize or extract utterances related to REMED, ChatGPT overestimated the mathematical instruction quality of the utterances and falsely identified relevant sentences that were leading questions regardless of whether they were addressing student errors and difficulty or whether they were at conceptual level.
It failed to make inferences based on the level of pedagogical expertise required for assessing these subject matter variables.
The finding is consistent with \citet{wang-demszky-2023-chatgpt}, who found that $30\%$ of the highlights retrieved by ChatGPT are unfaithful, $55\%$ are irrelevant, and $71\%$ are not insightful, determined by education experts. Such poor preliminary results, along with the high cost, warrant no further large-scale experiments.

\paragraph{Model choices.}
Among all the PLMs employed in the experiments, DistilBERT, RoBERTa, and XLNet have comparable performance, outperforming BERT in many cases.
Notably, fine-tuning Llama2 with parameter-efficient methods results in unsatisfactory results in these subject-matter tasks.
As shown in Tables~\ref{tab:simse_dev_three} and~\ref{tab:simse_dev_binary}, it only marginally improves the majority baseline in most cases.
We also experimented with using other parameter-efficient methods like p-tuning \cite{liu2023gpt} and training Llama2 by adding a classification head on top of the hidden layers.
The results are similar.
Therefore, we report the results of the commonly adopted setup of fine-tuning Llama2 as a next-word prediction task~\cite{touvron2023llama}, by converting the dataset to instruction-following (Appendix \ref{app:training}).

\section{Related Work}
Prior works on automatic teaching evaluations primarily focused on low-level instruction statistics such as the frequency
of certain utterance-level teaching strategies employed in the classroom. Considering teachers' questions in the classroom play a crucial role in students' engagement and academic achievement \cite{kelly2007classroom}, many researchers studied question-related behaviors such as detecting utterances that contain questions \cite{blanchard2016semi, donnelly2017words}, identifying authentic questions without prescripted answers \cite{cook2018open, kelly2018automatically, reilly2019predicting}, identifying focusing and funneling questions \citep{alic2022computationally}, and so on \citep{huang2020neural}.
Other discourse features of teachers have also been shown to influence learning achievement. For instance, \citet{hunkins2022beautiful} detects socio-emotional supports. \citet{demszky2021measuring} measures teachers’ uptake to provide feedback for students' contribution. \citet{schlotterbeck2021tarta, xu2020automatic} recognizes different types of teaching activities such as presenting, guiding, and management \citep{tan2023sit}.

Despite previous efforts on low-level evaluation, \citet{wang-demszky-2023-chatgpt} provided some preliminary results on evaluating teaching quality and providing high-level feedback, which aligns more with the current forms of observation feedback in education. However, their study focused on testing ChatGPT's zero-shot ability and only evaluated a small set of MQI variables in the NCTE dataset. 


\section{Future Work}

\paragraph{Model Choices.} 
Although our experiments cover a wide range of PLMs including the recently released Llama2-7B model, many advanced Large Language Models (LLMs) were beyond our scope due to resource demands. With the negative results in the zero-shot setting \cite{wang-demszky-2023-chatgpt}, future research could focus on adapting these models for similar tasks, i.e., make LLMs better follow the instructions in the rubrics and align them with experts' preferences.

\paragraph{Model Explainability.} 
Our study indicates that fine-tuned PLMs can match expert-level performance on some observation variables. However, they lack transparent explanations for their predictions. Explainable models would not only enhance trust among education practitioners~\cite{lipton2018mythos} but could also enrich the feedback that teachers receive. Our preliminary analysis using attribution score to highlight the important tokens achieves unsatisfactory results, suggesting more efforts on this venue. 
Additionally, PLMs often rely on spurious correlations~\cite{zhou2023explore, tu2020empirical}, underscoring the value of explainable models in uncovering prediction rationales and enhancing model robustness.
Another way of providing insights into models' behaviors is to generate free-text explanations, which seems promising with the recent advance of generative LLMs.

\paragraph{Multimodality.}
Education recordings inherently encompass multiple modalities, offering rich data through visuals like facial expressions and acoustics such as vocal tones, which are pivotal for education specialists in assessing teaching quality. 
Yet, these features are underexplored in automated methods due to the lack of high-quality video or audio data. Advances in recording technology and the rise of online educational platforms can facilitate the collection of high-quality audio-visual content, paving the road for future research.
Nonetheless, the sensitive nature of this data necessitates careful handling to prioritize privacy and ethical issues.

\section{Conclusion}

In this study, we present the most comprehensive study to date on assessing high-level instruction quality, spanning both in-person and simulated educational settings, with a wide coverage of observation variables for math education, including metacognitive components tailored for students with special needs. We find that the performance of PLMs largely depends on the level of pedagogical expertise required for assessing the observation variables, despite employing various techniques aimed at addressing the issues of lengthy inputs and the scarcity of high-rating samples. Our study enhances the understanding of the extent to which current NLP techniques work in education and brings insights into automatically evaluating high-inference instruction quality, which could reduce disparities in teachers' professional learning and, consequently, students' academic outcomes.

\clearpage
\section*{Limitations}
\paragraph{Generalizability.} Although our study includes two distinct educational settings, the NCTE datset is the largest available dataset of U.S. classroom, and the SimSE dataset focuses on students with special needs, they still only capture a small proportion of U.S. classrooms and teaching activities. Moreover, the NCTE dataset was collected over a decade ago and thus may not represent the current education field, even though teaching practices have remained relatively constant over time \cite{cuban1993teachers, cohen2017reform}. Therefore, there are limitations to the generalizability of the findings. It is important to carefully validate the measures built on these datasets before employing them in a new domain or a target population.

\paragraph{Lack of High-Rating Teaching Samples.}
One issue we tried to address in this study is the lack of high-rating teaching samples. However, this is limited by the lack of high-quality teaching activities in real-world classrooms/simulation sessions \cite{kane2012gathering}. In fact, the teaching simulation platform that curated the SimSE dataset aims to improve teachers' professional development, along with a lot of ongoing efforts to provide teachers with timely feedback \cite{demszky2023m}. Moreover, with the recent advance of generative LLMs~\cite{liu2024large}, researchers have started exploring the potential of using them to generate educational content \cite{li2023curriculum, zelikman-etal-2023-generating}. Despite their relatively poor performances in solving math problems \cite{frieder2024mathematical}, coaching on mathematical education \cite{wang-demszky-2023-chatgpt}, and remediating students' mathematical errors \cite{wang2023step}, the generated content can still be helpful to some extent.

\section*{Acknowledgements}
We thank Lindsey McLean and Marissa Pilger Suhr for curating and sharing the SimSE dataset. We thank Dorottya Demszky and Dan McGinn for their help with the NCTE dataset.
We thank the financial support from the National Science Foundation through award numbers 2009939 and 2010298, as well as the Grand Challenge program at the University of Maryland.
\bibliography{anthology,custom}

\appendix
\section{Additional MQI Variables}
\label{app:mqi_additional}
For Richness of the Mathematics, we have \textit{LINK} that measures students' and teachers' explicit linking and connections between different representations of a mathematical idea or procedure; \textit{EXPL} that evaluates the quality of the teacher’s mathematical explanations; \textit{MMETH} that evaluates the quality when multiple procedures or solution methods are discussed.
For Errors and Imprecision, we have 
\textit{MAJERR} that captures major mathematical errors and 
\textit{LCP} that captures when a teacher’s utterances cannot be understood.
For Student Participation in Meaning-Making and Reasoning, we have 
\textit{SMQR} that measures how students engage in mathematical thinking that has features of important mathematical practices; and \textit{ETCA} that captures student engagement in tasks in which they think deeply and reason about mathematics, regardless of the initial demand of the curriculum/textbook task or how the teacher sets up the task for students.

The statistics and results for these additional MQI variables are shown in Table \ref{tab:ncte_test_additional} and Table \ref{tab:ncte_test_transcript_additional}.

\begin{table}[!htb]
\centering
\resizebox{\linewidth}{!}{%
\begin{tabular}{lrrrrr} \toprule
\multirow{2}{*}{Variable} & \multicolumn{3}{c}{Three-way}      & \multicolumn{2}{c}{Binary}  \\
\cmidrule(lr){2-4}\cmidrule(l){5-6}
                      & Majority F1 & F1        & Spearman  & Majority F1 & F1            \\ \midrule
LINK                  & 25±0        & 53±1      & 0.46±0.02 & 37±0        & 71±1          \\
EXPL                  & 25±0        & 46±2      & 0.32±0.02 & 38±0        & 66±2          \\
MMETH                 & 29±0        & 51±3      & 0.39±0.03 & 44±0        & 70±1          \\
MAJERR                & 28±0        & 37±1      & 0.12±0.03 & 48±0        & 55±1          \\
LCP                   & 30±0        & 37±1      & 0.15±0.04 & 46±0        & 58±1          \\
SMQR                  & 27±0        & 45±2      & 0.26±0.02 & 41±0        & 62±1          \\
ETCA                  & 24±0        & 48±2      & 0.32±0.01 & 36±0        & 64±0          \\ \bottomrule
\end{tabular}
}
\caption{Test set performance of selected models for the additional seven MQI variables in the NCTE dataset. F1 scores are in percentage. Inputs use teacher text only.} 
\label{tab:ncte_test_additional}
\end{table}

\begin{table}[!htb]
\centering
\small
\begin{tabular}{lrrr} \toprule
\multirow{2}{*}{Variable} & \multicolumn{2}{c}{Three-way} & Binary     \\
\cmidrule(lr){2-3}\cmidrule(lr){4-4}
                      & F1 (\%)       & Spearman           & F1 (\%)         \\ \midrule
LINK                  & 50±1      & 0.41±0.02          & 70±1       \\
EXPL                  & 46±1      & 0.31±0.03          & 65±2       \\
MMETH                 & 50±3      & 0.36±0.02          & 68±1       \\
MAJERR                & 37±0      & 0.13±0.01          & 56±1       \\
LCP                   & 37±1      & 0.13±0.02          & 57±1       \\
SMQR                  & 45±0      & 0.26±0.03          & 62±1       \\
ETCA                  & 46±1      & 0.30±0.02          & 65±0       \\ \bottomrule
\end{tabular}
\caption{Test set performance of selected models for the additional seven MQI variables in the NCTE dataset. Inputs are in the transcript format.}
\label{tab:ncte_test_transcript_additional}
\end{table}

\section{Computation for Spearman Correlation among Human Raters}
\label{app:rater}
For the SimSE dataset, we describe the computation process for the Spearman correlation among human raters. For each pair of raters, we calculate the Spearman correlation on the samples that they both annotated, as shown in Table~\ref{tab:rater}. The final correlation values in Table~\ref{tab:simse_test} are weighted average values weighted by the sample size for each pair of raters.

\begin{table}[!htb]
\centering
\resizebox{\linewidth}{!}{%
\begin{tabular}{cccccccc} \toprule
Rater1 & Rater2 & $N$ & Objective & Unpacking & Self-Instruction & Self-Regulation & Ending  \\ \hline
1      & 2      & 80          & 0.87      & 0.81      & 0.33             & 0.52            & 0.88    \\
1      & 3      & 28          & 0.56      & 0.55      & -0.08            & 0.85            & 0.87    \\
1      & 4      & 76          & 0.69      & 0.45      & 0.36             & 0.53            & 0.76    \\
1      & 5      & 44          & 0.95      & 0.55      & 0.34             & 0.45            & 0.73    \\
1      & 6      & 51          & 0.77      & 0.64      & 0.44             & 0.56            & 0.78    \\
1      & 7      & 46          & 0.89      & 0.68      & 0.64             & 0.82            & 0.87    \\
1      & 8      & 41          & 0.82      & 0.42      & 0.68             & 0.77            & 0.77    \\
2      & 3      & 30          & 0.91      & 0.88      & 0.86             & 0.86            & 0.66    \\
2      & 4      & 54          & 0.87      & 0.60      & 0.44             & 0.68            & 0.52    \\
2      & 5      & 41          & 0.70      & 0.30      & 0.38             & 0.47            & 0.58    \\
2      & 6      & 61          & 0.77      & 0.69      & 0.19             & 0.66            & 0.92    \\
2      & 7      & 49          & 0.64      & 0.51      & 0.36             & 0.55            & 0.90    \\
2      & 8      & 56          & 0.72      & 0.48      & 0.73             & 0.69            & 0.77    \\
3      & 4      & 23          & 0.61      & 0.53      & -                & 0.59            & 0.76    \\
3      & 5      & 20          & 0.79      & 0.80      & -                & 1.00            & 0.44    \\
3      & 6      & 22          & 0.81      & 0.51      & 0.69             & 0.69            & 0.87    \\
3      & 7      & 32          & 0.77      & 0.31      & -0.12            & 0.76            & 0.70    \\
4      & 5      & 47          & 0.80      & 0.49      & -0.06            & 0.86            & 0.82    \\
4      & 6      & 58          & 0.82      & 0.50      & 0.33             & 0.47            & 0.62    \\
4      & 7      & 55          & 0.78      & 0.56      & 0.20             & 0.86            & 0.87    \\
4      & 8      & 32          & 0.77      & 0.68      & 0.40             & 0.56            & 0.74    \\
5      & 6      & 41          & 0.81      & 0.39      & -0.06            & -               & 0.59    \\
5      & 7      & 58          & 0.77      & 0.70      & 0.29             & 0.69            & 0.85    \\
5      & 8      & 34          & 0.79      & 0.26      & 0.51             & 0.39            & 0.85    \\
6      & 7      & 53          & 0.83      & 0.68      & 0.55             & 0.86            & 0.97    \\
6      & 8      & 24          & 0.91      & 0.52      & 0.34             & -               & 0.91    \\
7      & 8      & 23          & 0.90      & 0.91      & -0.07            & 0.69            & 0.94    \\ \bottomrule
\end{tabular}
}
\caption{Spearman correlation between each pair of raters in the SimSE dataset. $N$ denotes sample size and $-$ indicates there is only one type of rating in the samples that two raters annotated for the component, resulting in invalid correlation values.}
\label{tab:rater}
\end{table}

\begin{table*}[!htb]
\centering
\resizebox{\linewidth}{!}{%
\begin{tabular}{lccccccccccccccccc} \toprule
\multirow{2}{*}{Component} & \multicolumn{5}{c}{Normal}                   & \multicolumn{4}{c}{Weighted}        & \multicolumn{4}{c}{Two-stage}       & \multicolumn{4}{c}{Two-stage + Weighted}  \\
\cmidrule(lr){2-6}\cmidrule(lr){7-10}\cmidrule(lr){11-14}\cmidrule(lr){15-18}
                           & distilbert & bert  & xlnet & roberta & llama2 & distilbert & bert  & xlnet & roberta & distilbert & bert  & xlnet & roberta & distilbert & bert & xlnet & roberta       \\ \midrule
Objective                  & \textbf{89±2} & 79±22 & 70±27 & 75±23   & 63±7   & 89±3          & 88±4  & 74±22         & 89±3    & 88±2          & 88±2  & 88±2  & 86±4    & 88±2          & 88±2 & 83±6  & 88±2       \\
Unpacking                  & \textbf{77±1} & 61±23 & 66±18 & 76±2    & 53±16  & \textbf{77±1} & 65±19 & 73±3          & 76±1    & 74±1          & 67±18 & 59±22 & 75±2    & 74±2          & 75±1 & 67±19 & 67±17      \\
Self-Instruction           & 75±2          & 73±7  & 56±18 & 68±15   & 49±9   & 78±1          & 59±17 & 68±14         & 70±15   & 76±2          & 77±3  & 71±16 & 60±16   & \textbf{77±2} & 75±4 & 70±15 & 73±5       \\
Self-Regulation            & 81±3          & 71±15 & 69±21 & 76±18   & 48±5   & 81±5          & 59±18 & 56±14         & 46±0    & \textbf{86±2} & 85±5  & 80±3  & 75±16   & 86±4          & 85±2 & 81±4  & 81±4       \\
Ending                     & 86±2          & 79±20 & 84±23 & 78±20   & 69±10  & 86±2          & 82±8  & \textbf{94±2} & 87±3    & 82±7          & 79±11 & 70±18 & 76±12   & 83±4          & 83±4 & 80±6  & 82±8 \\\bottomrule
\end{tabular}
}
\caption{The F1 scores (\%) of predicting metacognitive modeling components performance on Dev set of SimSE dataset in the binary setting. The best-performing models are bolded.}
\label{tab:simse_dev_binary}
\end{table*}

\section{Training Details}
\label{app:training}
For BERT, DistilBERT, XLNet, and RoBERTa, the model checkpoints are bert-large-uncased, distilbert-base-uncased, xlnet-large-cased, and roberta-large. we used a fixed set of hyperparameters, where the learning rate is $2\times 10^{-5}$ and batch size is $4$. Each model is trained for $6$ epochs and the checkpoint with the best F1 performance on the Dev set is selected for comparisons with other models. 

For Llama2, we experimented with $7$B and $13$B model versions. However, we did not observe satisfying performances with these models after several trials with different training paradigms (i.e., QLoRA and P-tuning) and hyperparameter settings. For QLora, We took inspiration from \cite{lorafinetune} and experimented $\{r=8, \alpha=16\}$, $\{r=16, \alpha=32\}$, $\{r=128, \alpha=256\}$, $\{r=256, \alpha=512\}$, and $\{r=512, \alpha=1024\}$.
Here we report the hyperparameters for the results with QLoRA as shown in Table~\ref{tab:ncte_dev_three} and Table~\ref{tab:ncte_dev_binary}. The hyperparameters for QLoRA are $r=256, \alpha=512$ with a dropout rate of $0.1$. The learning rate for training is $2\times 10^{-4}$, batch size is $2$ with gradient accumulation step $4$, trained for $5$ epochs. 
Each training sample is converted to the following instruction-following prompt: \texttt{Transcript: \{transcript\} Rating: \{rating\}} and the prompt for test samples is \texttt{Transcript: \{transcript\} Rating:}. 
All models are trained on one RTXA6000 GPU.

\section{Additional Dev Set Performance}
\label{app:dev}

This section shows the Dev set performances for the metacognitive modeling components in the SimSE dataset in the binary setting (Table~\ref{tab:simse_dev_binary}), and for the MQI variables from the NCTE dataset, in the binary (Table~\ref{tab:ncte_dev_three}) and three-way (Table~\ref{tab:ncte_dev_binary}) settings.

\begin{table}[!htb]
\centering
\resizebox{\linewidth}{!}{%
\begin{tabular}{lrrrrrr} \toprule
\multirow{2}{*}{Variable} & \multicolumn{3}{c}{Normal}                                  & \multicolumn{3}{c}{Weighted}                \\
\cmidrule(lr){2-4}\cmidrule(lr){5-7}
                          & distilbert         & bert               & roberta            & distilbert         & bert      & roberta    \\ \midrule
LINK                      & 48±3               & 46±12              & \textbf{53±1}      & 51±1               & 41±15     & 32±12      \\
EXPL                      & 44±0               & 42±10              & 46±3               & \textbf{47±2}      & 30±10     & 36±10      \\
MMETH                     & 45±1               & 45±9               & 45±9               & \textbf{51±2}      & 47±10     & 47±10      \\
MLANG                     & 50±2               & \textbf{54±1}      & 53±1               & \textbf{54±1}      & 36±18     & 39±16      \\
REMED                     & 41±0               & 37±10              & 45±2               & \textbf{45±1}      & 29±10     & 34±11      \\
MAJERR                    & 38±2               & 35±3               & 35±3               & \textbf{39±2}      & 36±3      & 34±3       \\
LANGIMP                   & 40±1               & 40±1               & 38±5               & \textbf{45±1}      & 40±7      & 40±7       \\
LCP                       & \textbf{38±1}      & \textbf{38±1}      & 34±4               & 37±1               & 35±5      & 37±4       \\
STEXPL                    & 45±1               & \textbf{46±1}      & 45±1               & \textbf{46±1}      & 44±9      & 40±9       \\
SMQR                      & 40±0               & 41±2               & 37±9               & \textbf{44±2}      & 35±8      & 37±7       \\
ETCA                      & 43±1               & 42±11              & 47±4               & \textbf{47±1}      & 32±11     & 34±14      \\ \bottomrule
\end{tabular}
}
\caption{F1 scores (\%) on the NCTE Dev dataset in the three-way setting. }
\label{tab:ncte_dev_three}
\end{table}

\begin{table}[!htb]
\centering
\resizebox{\linewidth}{!}{%
\begin{tabular}{lrrrrrr} \toprule
\multirow{2}{*}{Variable} & \multicolumn{3}{c}{Normal}                                  & \multicolumn{3}{c}{Weighted}                \\
\cmidrule(lr){2-4}\cmidrule(lr){5-7}
                          & distilbert         & bert               & roberta            & distilbert         & bert      & roberta    \\ \midrule
LINK                      & \textbf{70±1}      & 63±15              & 57±18              & \textbf{70±1}      & 57±18     & 65±15      \\
EXPL                      & \textbf{67±2}      & 54±15              & 49±16              & 66±1               & 60±13     & 59±13      \\
MMETH                     & 69±2               & 60±15              & 69±4               & \textbf{69±1}      & 59±14     & 57±13      \\
MLANG                     & \textbf{70±0}      & 64±15              & 70±1               & 70±1               & 70±1      & 63±15      \\
REMED                     & 64±1               & \textbf{65±1}      & 53±17              & 64±1               & 53±16     & 53±15      \\
MAJERR                    & 57±2               & 49±4               & 52±4               & \textbf{57±1}      & 48±1      & 49±4       \\
LANGIMP                   & 59±1               & 57±8               & 58±8               & \textbf{61±1}      & 51±10     & 54±10      \\
LCP                       & 57±2               & 52±6               & 48±5               & \textbf{58±1}      & 51±6      & 50±7       \\
STEXPL                    & 68±1               & 69±1               & \textbf{70±1}      & 68±1               & 63±12     & 54±14      \\
SMQR                      & \textbf{62±1}      & 46±9               & 46±10              & \textbf{62±1}      & 49±11     & 58±9       \\
ETCA                      & 64±1               & \textbf{65±1}      & 54±16              & 64±1               & 42±14     & 36±1       \\ \bottomrule
\end{tabular}
}
\caption{F1 scores on the NCTE Dev dataset in the binary setting.}
\label{tab:ncte_dev_binary}
\end{table}

\section{Qualitative Analysis of Summarization with ChatGPT}
\begin{figure}
    \centering
    \includegraphics[width=0.90\linewidth]{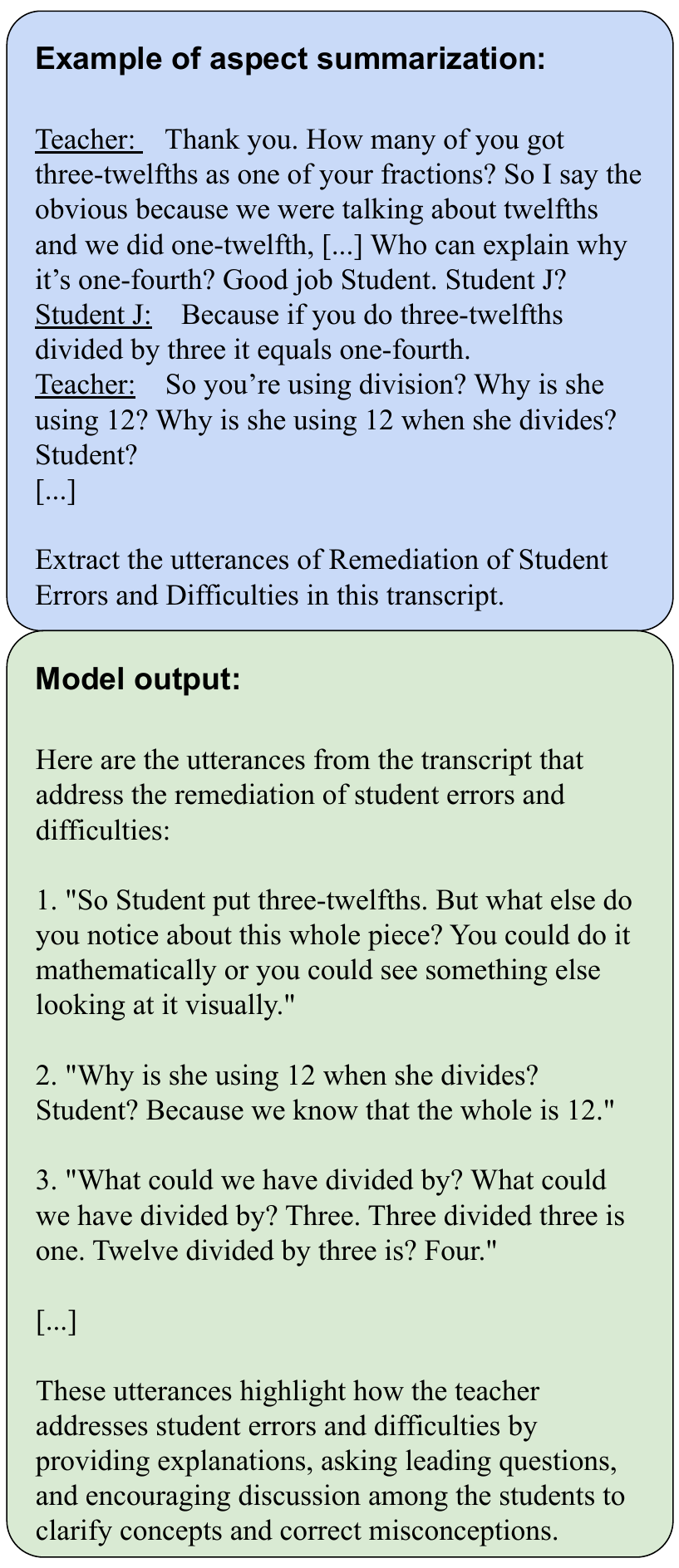}
    \caption{An example for using ChatGPT to extract utterances relevant to REMED. The example is rated as low in the NCTE dataset, while ChatGPT overestimates the quality of the sample.}
    \label{fig:prompt}
\end{figure}

Figure~\ref{fig:prompt} shows a qualitative example of using ChatGPT with a summarization prompt in the first stage of the proposed two-stage prediction strategy.

\end{document}